\documentclass[twoside,11pt]{article}

\usepackage[preprint]{jmlr2e}

\usepackage[utf8]{inputenc}
\usepackage[T1]{fontenc}
\usepackage{amsmath}
\usepackage{amssymb}
\usepackage{commath}
\usepackage{graphicx}
\usepackage{subcaption}
\usepackage{cleveref}

\newcommand{\beginsupplement}{%
        \setcounter{table}{0}
        \renewcommand{\thetable}{S\arabic{table}}%
        \setcounter{figure}{0}
        \renewcommand{\thefigure}{S\arabic{figure}}%
     }

\usepackage{lastpage}

\jmlrheading{?}{2023}{1-\pageref{LastPage}}{10/23; Revised ?/?}{?/?}{papr-id}{Dimitrije Markovi\'c, Karl J. Friston, and Stefan J. Kiebel}

\ShortHeadings{Sparsification with Bayesian model reduction}{Markovi\'c, Friston, and Kiebel}
\firstpageno{1}

\begin{document}

\title{Bayesian sparsification for deep neural networks with Bayesian model reduction}

\author{\name Dimitrije Markovi\'c$^{1, \dag}$ \email dimitrije.markovic@tu-dresden.de \\
       \AND
       \name Karl J. Friston$^{3, 4}$ \email k.friston@ucl.ac.uk \\
       \AND
       \name Stefan J. Kiebel$^{1, 2}$ \email stefan.kiebel@tu-dresden.de \\
       \addr 
       $^{1}$ Chair of Cognitive Computational Neuroscience, Technische Universit\"at Dresden, Dresden, Germany\\
       $^{2}$ Centre for Tactile Internet with Human-in-the-Loop (CeTI), Technische Universit\"at Dresden, Dresden, Germany\\
       $^{3}$ Wellcome Centre for Human Neuroimaging, Queen Square Institute of Neurology, University College London, London, UK\\
       $^{4}$ VERSES AI Research Lab, Los Angeles, California, 90016, USA.\\
       $^{\dag}$ Corresponding author}

\editor{?}

\maketitle

\begin{abstract}%
Deep learning's immense capabilities are often constrained by the complexity of its models, leading to an increasing demand for effective sparsification techniques. Bayesian sparsification for deep learning emerges as a crucial approach, facilitating the design of models that are both computationally efficient and competitive in terms of performance across various deep learning applications. The state-of-the-art -- in Bayesian sparsification of deep neural networks -- combines structural shrinkage priors on model weights with an approximate inference scheme based on stochastic variational inference. However, model inversion of the full generative model is exceptionally computationally demanding, especially when compared to standard deep learning of point estimates. In this context, we advocate for the use of Bayesian model reduction (BMR) as a more efficient alternative for pruning of model weights. As a generalization of the Savage-Dickey ratio, BMR allows a post-hoc elimination of redundant model weights based on the posterior estimates under a straightforward (non-hierarchical) generative model. Our comparative study highlights the advantages of the BMR method relative to established approaches based on hierarchical horseshoe priors over model weights. We illustrate the potential of BMR across various deep learning architectures, from classical networks like LeNet to modern frameworks such as Vision Transformers and MLP-Mixers.
\end{abstract}


\section{Introduction}

Bayesian deep learning integrates the principles of Bayesian methodology with the objectives of deep learning, facilitating the training of expansive parametric models tailored for classifying and generating intricate audio-visual data, including images, text, and speech \citep{wang2020survey, wilson2020a,wang2016towards}. Notably, the Bayesian approach frames the challenge of model optimization as an inference problem. This perspective is especially apt for scenarios necessitating decision-making under uncertainty \citep{murphy2022probabilistic,ghahramani2015probabilistic}. As a result, Bayesian formulations in deep learning have proven advantageous in various respects, offering enhancements in generalization \citep{wilson2020b}, accuracy, calibration \citep{izmailov2020subspace,luo2020}, and model compression \citep{louizos2017bayesian}.

These functional enhancements are intrinsically tied to judiciously chosen structural priors \citep{fortuin2022priors}. The priors, integral to the probabilistic generative model, scaffold the architecture of the network, thereby reducing the data required for the inference of optimal parametric solutions. Recent studies have highlighted the efficacy of hierarchical shrinkage priors over model weights, a specific category of structural priors, in achieving highly-sparse network representations \citep{nalisnick2019dropout,louizos2017bayesian,seto2021,ghosh2018structured}. Sparse representations not only reduce redundancy but also evince additional performance benefits. However, the adoption of shrinkage priors in all deep learning models presents a conundrum: the ballooning space of latent parameters and the diminishing scalability of prevailing approximate inference schemes \citep{snoek2015scalable,krishnan2019,izmailov2020subspace,daxberger2021}.

In line with ongoing research on scalable Bayesian inference, we introduce an approximate inference scheme rooted in Bayesian model reduction (BMR). In essence, BMR extends the foundational principles of the Savage-Dickey Density Ratio method \citep{cameron2013}. BMR is typically conceptualized as a combinatorial model comparison framework, enabling swift estimations of model evidence across an extensive array of models, that differ in their prior assumptions, to identify the most probable one. Originally conceived for model comparison within the dynamical causal modeling framework \citep{rosa2012post,friston2011post}, the scope of BMR has since broadened. Subsequent works expanded its methodology \citep{friston2016bayesian,friston2017active,friston2018bayesian} and adapted it for structure learning \citep{smith2020active}. More recently, BMR has found applications in Bayesian nonlinear regression and classification tasks using Bayesian neural networks with variance backpropagation \citep{beckers2022principled,haussmann2020sampling}.

The BMR method is intimately connected with the spike-and-slab prior, a type of shrinkage prior \citep{mitchell1988bayesian}. Intriguingly, this specific structured shrinkage prior has parallels with Dropout regularization \citep{nalisnick2019dropout}. Such an association spurred researchers in Bayesian deep learning to formulate sparsification methods based on a different type of shrinkage prior---the hierarchical horseshoe prior \citep{piironen2017sparsity}---as a tool for automated depth determination. Subsequent studies suggested that merging horseshoe priors with structured variational approximations yields robust, highly sparse representations \citep{ghosh2018structured}. The allure of continuous shrinkage priors (e.g., horseshoe priors) stems from the computational challenges associated with model inversion reliant on spike-and-slab priors \citep{nalisnick2019dropout,piironen2017sparsity}. However, continuous shrinkage priors necessitate a considerably more expansive parameter space, to represent the approximate posterior, compared to optimizing neural networks using the traditional point estimate method.

In this work, we reexamine the spike-and-slab prior within the framework of BMR-based sparsification, highlighting its efficiency. Notably, this approach circumvents the need to expand the approximate posterior beyond the conventional fully factorised mean-field approximation, making it more scalable than structured variational approximations \citep{ghosh2018structured}. In this light, BMR can be seen as a layered stochastic and black-box variational inference technique, which we term \emph{stochastic BMR}. We subject the stochastic BMR to rigorous validation across various image classification tasks and network architectures, including LeNet-5 \citep{lecun1989backpropagation}, Vision Transformers \citep{dosovitskiy2020image}, and MLP-Mixers \citep{tolstikhin2021mlp}.

Central to our study is an empirical comparison of stochastic BMR with methods anchored in hierarchical horseshoe priors. Through multiple metrics - from Top-1 accuracy to expected calibration error and negative log-likelihood - we establish the competitive performance of stochastic BMR. We argue its computational efficiency, and remarkable sparsification rate, position BMR as an appealing choice, enhancing the scalability and proficiency of contemporary deep learning networks across diverse machine learning challenges, extending well beyond computer vision. We conclude with a discussion on potential avenues of future research that could further facilitate of BMR based pruning of deep neural networks.

\section{Methods}

In this section, we first describe the methods and techniques used in our research to address the problem of efficient Bayesian sparsification of deep neural networks. We provide a detailed overview of our approach, starting with variational inference methods, followed by the formulation of the Bayesian model reduction (BMR), Bayesian neural networks with shrinkage priors, and the description of corresponding approximate posterior.

\subsection{Variational inference}
Given a joint density of latent variables, represented as $\pmb{z} = \del{z_1, \ldots, z_k}$, and a dataset of $n$ observations $\pmb{\mathcal{D}} = \del{y_1, \ldots, y_n}$ we can express the joint density, that is, the generative model, as
\begin{equation*}
    p\del{\pmb{\mathcal{D}}, \pmb{z}} = p\del{\pmb{z}} p\del{\pmb{\mathcal{D}}|\pmb{z}}.
\end{equation*}
The posterior density is then obtained, following the Bayes rule, as
\begin{equation}
    p\del{\pmb{z}|\pmb{\mathcal{D}}} \propto p\del{\pmb{z}} p\del{\pmb{\mathcal{D}}|\pmb{z}}.
    \label{eq:exact}
\end{equation}

For complex generative models, direct inference as described above becomes computationally prohibitive. To circumvent this, we approximate the exact posterior $p\del{\pmb{z}|\pmb{\mathcal{D}}}$, constraining it to a distribution $q\del{z}$ that belongs to a named distribution family $\mathcal{Q}$. We then seek $q^*\del{z} \in \mathcal{Q}$, an approximate solution that minimizes the following Kullback-Leibler divergence \citep{blei2017variational}
\begin{equation*}
    q^*\del{z} = \underset{q \in \mathcal{Q}}{\text{argmin} } D_{KL} \del{q\del{\pmb{z}}||p\del{\pmb{z}|\pmb{\mathcal{D}}}} = \underset{q \in \mathcal{Q}}{\text{argmin} } F \sbr{q}, 
\end{equation*}
where $F\sbr{q}$ stands for the variational free energy (VFE), defined as 
\begin{equation*}
    F\sbr{q} = E_{q\del{\pmb{z}}} \sbr{\ln q \del{\pmb{z}} - \ln p\del{\pmb{\mathcal{D}}, \pmb{z}}}
\end{equation*}
VFE serves as an upper bound on the marginal log-likelihood
$$ F\sbr{q} = D_{KL} \del{q\del{\pmb{z}}||p\del{\pmb{z}|\pmb{\mathcal{D}}}} - \ln p\del{\pmb{\mathcal{D}}} \geq - \ln p\del{\pmb{\mathcal{D}}} $$
As KL-divergence is always greater or equal to zero, minimizing VFE brings the approximate solution as close as possible to the true posterior, without having to compute the exact posterior.

The most straightforward way to obtain the approximate posterior $q^*\del{\pmb{z}}$, is to minimize the VFE along its negative gradient:
$$ \dot{\pmb{\phi}} = - \nabla_{\pmb{\phi}} F\sbr{q} $$
where $\pmb{\phi}$ signifies the parameters of the approximate posterior $q_{\pmb{\phi}}\del{\pmb{z}} = q\del{\pmb{z}|\pmb{\phi}}$. Thus, variational inference reframes the inference problem highlighted in \cref{eq:exact} as an optimization problem \cite{beal2003variational}.

\subsection{Stochastic and black-box variational inference}

\emph{Stochastic variational inference} (SVI) improves the computational efficiency of gradient descent by approximating the variational free energy using a subset---$\pmb{\mathcal{K}}_i = \del{y_{s^i_1}, \ldots, y_{s^i_k}}; \:\: k \ll n$---of the entire data set $\pmb{\mathcal{D}}$. This approach fosters a stochastic gradient descent (SGD) mechanism, capable of managing large datasets \citep{hoffman2013stochastic}. Crucially, at every iteration step $i$ of the SGD process, the subset $\pmb{\mathcal{K}}_i$ undergoes re-sampling.

\emph{Black-box Variational Inference} (BBVI) facilitates the optimization of any (named or unnamed) posterior density $q_{\pmb{\phi}}\del{\pmb{z}}$, through the integration of Monte Carlo estimates for variational gradients \citep{ranganath2014black}. This can be formulated as the following relation
\begin{equation}
    \nabla_{\pmb{\phi}} F\sbr{q} \approx \nabla_{\pmb{\phi}} \hat{F}\sbr{q} = \frac{1}{S}\sum_{s=1}^S \nabla_{\pmb{\phi}} \ln q_{\pmb{\phi}}\del{\pmb{z}} \sbr{ \ln \frac{q_{\pmb{\phi}}\del{\pmb{z}}}{p\del{\pmb{\mathcal{D}}, \pmb{z}}} + 1}; \quad \pmb{z}_s \sim q\del{\pmb{z}|\pmb{\phi}} 
    \label{eq:reinforce}
\end{equation}
which is known as the REINFORCE estimator \citep{williams1992simple}. To mitigate the variance inherent to Monte Carlo gradient estimations, we employ Rao-Blackwellization \citep{schulman2015gradient}, with an implementation sourced from NumPyro \citep{bingham2019pyro}. For optimizing the variational objective stochastically, we leverage the AdaBelief optimizer \citep{zhuang2020adabelief}. As an adaptive algorithm, AdaBelief ensures swift convergence, robust generalization, and steady optimization. Notably, we use AdaBelief's implementation from the Optax package within the JAX ecosystem \citep{deepmind2020jax}.

\subsection{Bayesian model reduction}
Let us consider two generative processes for the data: a full model
\begin{equation*}
    p\del{\pmb{z}|\pmb{\mathcal{D}}} \propto p\del{\pmb{\mathcal{D}}| \pmb{z}} p\del{\pmb{z}}
\end{equation*} 
and a reduced model \footnote{the reduction here implies applying constraints of any form to the prior to obtain a posterior with reduced entropy.} in which the original prior $p\del{\pmb{z}}$ is replaced with a more informative prior $\tilde{p}\del{\pmb{z}} = p\del{\pmb{z}|\pmb{\theta}}$ that depends on hyper-parameters $\pmb{\theta}$. This change leads to a different posterior
\begin{equation*}
    \tilde{p}\del{\pmb{z}|\pmb{\mathcal{D}}} \propto p\del{\pmb{\mathcal{D}}| \pmb{z}} \tilde{p}\del{\pmb{z}}
\end{equation*}

Noting that as the following relation holds:
$$ 1 = \int \dif \pmb{z} \tilde{p}\del{\pmb{z}|\pmb{\mathcal{D}}} = \int \dif \pmb{z} p\del{\pmb{z}|\pmb{\mathcal{D}}} \frac{\tilde{p}\del{\pmb{z}} p\del{\pmb{\mathcal{D}}}}{p\del{\pmb{z}}\tilde{p}\del{\pmb{\mathcal{D}}}},$$
we can express the link between the models as:
\begin{equation}
    \begin{split}
   - \ln \tilde{p}\del{\pmb{\mathcal{D}}} &= - \ln p\del{\pmb{\mathcal{D}}} - \ln \int d \pmb{z} p\del{\pmb{z}|\pmb{\mathcal{D}}}\frac{\tilde{p}\del{\pmb{z}}}{p\del{\pmb{z}}} \\
   &\approx F\del{\pmb{\phi}^*} - \ln \int d \pmb{z} q_{\pmb{\phi}^*}\del{\pmb{z}}\frac{\tilde{p}\del{\pmb{z}}}{p\del{\pmb{z}}}
   \end{split}
   \label{eq:link}
\end{equation}
where we assumed the approximate posterior for the full model corresponds to $p\del{\pmb{z}|\pmb{\mathcal{D}}} \approx q_{\pmb{\phi}^*}\del{\pmb{z}}$, and that $ - \ln p\del{\pmb{\mathcal{D}}} \approx F\del{\pmb{\phi}^*}$.

From \cref{eq:link} we obtain the free energy of the reduced model as
\begin{equation}
    - \ln \tilde{p}\del{\pmb{\mathcal{D}}} \approx - \ln E_q\left[\frac{\tilde{p}\del{\pmb{z}}}{p\del{\pmb{z}}}\right] + F\del{\pmb{\phi}^*} = - \Delta F\del{\pmb{\theta}}.
    \label{eq:loglkl}
\end{equation}
where $\Delta F\del{\pmb{\theta}}$ denotes the change in the free energy of going from the full model to the reduced model, given hyper-parameters $\pmb{z}_H$. Note that for $\Delta F\del{\pmb{\theta}} > 0$ the reduced model has a better variational free energy compared to the flat model. Consequently, the reduced model offers a model with a greater marginal likelihood; i.e., a better explanation for the data and improved generalization capabilities. Heuristically, this can be understood as minimising model complexity, without sacrificing accuracy (because log evidence can be expressed as accuracy minus complexity, where complexity is the KL divergence between posterior and prior beliefs). This relationship is pivotal in formulating efficient pruning criteria, especially for extensive parametric models commonly employed in deep learning. 

\subsection{Bayesian neural networks}

In a general (nonlinear) regression problem, we model the relationship between predictors $\pmb{X} = \del{\pmb{x}_1, \ldots, \pmb{x}_n}$ and target variables $\pmb{Y} = \del{\pmb{y}_1, \ldots, \pmb{y}_n}$ using a likelihood distribution from an exponential family as
\begin{equation}
    \pmb{y}_i \sim p\del{\pmb{y}|\pmb{\mathcal{W}}, \pmb{x}_i} = h(y)\exp\sbr{\pmb{\eta}\del{\pmb{f}\del{\pmb{\mathcal{W}}, \pmb{x}_i}} \cdot \pmb{T}\del{\pmb{y}} - A\del{\pmb{f}\del{\pmb{\mathcal{W}}, \pmb{x}_i}}}.
    \label{eq:likelihood}
\end{equation}
Functions $h(\cdot), \pmb{\eta}(\cdot), \pmb{T}(\cdot), A(\cdot)$ are known and selected depending on the task. For example in a regression problem the likelihood will correspond to a multivariate normal distribution and in a classification problem to a categorical distribution. In this work, we will only consider a categorical likelihood, as it is the most suitable for image classification tasks. 

The mapping $\pmb{f}\del{\pmb{\mathcal{W}}, \pmb{x}_i}$ represents a generic deep neural network of depth $L$ defined as 
\begin{equation*}
\begin{split}
    \pmb{\mathcal{W}} &= \del{\pmb{W}_1, \ldots, \pmb{W}_L} \\
    \pmb{h}^0_i &= \pmb{x}_i \\
    \pmb{h}^l_i &= \pmb{g}\left( \pmb{W}_l \cdot \left[\pmb{h}_i^{l-1}; 1 \right] \right) \\
    \pmb{f}\del{\pmb{\mathcal{W}}, \pmb{x}_i} &= \pmb{W}_L \cdot \left[ \pmb{h}^{L-1}_i; 1\right]
\end{split}
\end{equation*}

A probabilistic formulation of the deep learning task, that is, inferring model weights, introduces implicit bias to the parameters $\pmb{\mathcal{W}}$ of an artificial neural network in the form of a prior distribution $p\del{\pmb{\mathcal{W}}}$. Hence, parameter estimation is cast as an inference problem where 
\begin{equation*}
    p\del{\pmb{\mathcal{W}}|\pmb{\mathcal{D}}} \propto p\del{\pmb{\mathcal{W}}} \prod_{i=1}^n p\del{\pmb{y}_i|\pmb{\mathcal{W}}, \pmb{x}_i} 
\end{equation*}

The choice of the prior distribution is crucial for optimal task performance, and a prior assumption of structural sparsity is essential for inferring sparse representations of over-parameterised models, such as deep neural networks. 

\subsection{Bayesian neural networks with shrinkage priors}
Shrinkage priors instantiate a prior belief about the sparse structure of model parameters. Here, we will investigate two well-established forms of shrinkage priors for network weight parameters, a canonical spike-and-slab prior \citep{george1993variable,mitchell1988bayesian} defined as 
\begin{equation*}
\begin{split}
        w_{ijl} & \sim \mathcal{N}\del{0, \lambda_{ijl}^2 \gamma_0^2} \\
        \lambda_{ijl} & \sim \text{Bernoulli}\del{\pi_{l}} \\
        \pi_l & \sim \mathcal{B}e\del{\alpha_0, \beta_0}
\end{split}
\end{equation*}

and a regularised-horseshoe prior \citep{piironen2017sparsity}
\begin{equation}
    \begin{split}
        w_{ijl} & \sim \mathcal{N}\del{0, \gamma_{il}^2} \\
        \gamma_{il}^2 &= \frac{c_l^2 v_l^2 \tau_{il}^2}{c_l^2 + \tau_{il}^2 v_l^2}\\
        c_l^{-2} &\sim \Gamma\del{2, 6} \\
        \tau_{il} &\sim \mathcal{C}^+(0, 1) \\
        v_l &\sim \mathcal{C}^+(0, \tau_0)
    \end{split}
    \label{eq:rhsprior}
\end{equation}
where $i \in \cbr{1, \ldots, K_l}$, $j \in \sbr{1, \ldots, K_{l-1}+1}$, and where $w_{ijl}$ denotes $ij$th element of the weight matrix at depth $l$. The symbols $\mathcal{B}e$, and $\mathcal{C}^+$ denote a Beta distribution and a half-Cauchy distribution, respectively.

Importantly, the spike-and-slab prior relates to dropout regularisation, which is commonly introduced as a sparsification method in deep learning \citep{nalisnick2019dropout,mobiny2021dropconnect}. This type of prior is considered the gold standard in shrinkage priors and has been used in many recent applications of Bayesian sparsification on neuronal networks \citep{bai2020efficient,hubin2023variational,jantre2021layer,sun2022learning,ke2022optimization} showing excellent sparsification rates. However, the inversion of the resulting hierarchical model is challenging and requires carefully constructed posterior approximations. Moreover, their dependence on discrete random variables renders them unsuitable for Markov-Chain Monte Carlo-based sampling schemes. As a result, researchers often use continuous formulations of the shrinkage-prior, with the horseshoe prior being a notable example.

In contexts that involve sparse learning with scant data, the regularised horseshoe prior has emerged as one of the preferred choices within shrinkage prior families \citep{ghosh2019model}. A distinct advantage of this prior is its ability to define both the magnitude of regularisation for prominent coefficients and convey information about sparsity. It is worth noting a dependency highlighted in \cite{ghosh2018structured}: for $v_l \tau_{il}  \ll 1$ the equation simplifies to $\gamma_{il} \approx  v_l \tau_{il}$ recovering the original horseshoe prior. In contrast, for $v_l \tau_{il} \gg 1$, the equation becomes $\gamma_{il}^2 \approx c_l^2$. In this latter scenario, the prior over the weights is defined as $w_{ijl} \sim \mathcal{N}\del{0, c^{2}_l}$, with $c_l$ serving as a weight decay hyper-parameter for layer $l$.

\subsection{Approximate posterior for Bayesian neural networks}

To benchmark stochastic BMR, we explore two forms of prior distribution $p\del{\pmb{\mathcal{W}}}$---a flat and a hierarchical structure---in conjunction with a fully factorised mean-field approximation.

Firstly, let us consider the flat prior over model weights, represented in a non-centered parameterization: 
\begin{equation}
\begin{split}
    c_l^{-2} &\sim \Gamma(2, 2) \\
    \hat{w}_{ijl} &\sim \mathcal{N} \del{0, 1} \\
    w_{ijl} &= \gamma_0 c_l \hat{w}_{ijl}
\end{split}
\label{eq:flatprior}
\end{equation}
where we set $\gamma_0 = 0.1$. Note that in the flat prior we incorporate a layer specific scale parameter, which we found to stabilise variational inference. Based on this, we describe a fully factorised approximate posterior as a composite of Normal and Log-Normal distributed random variables. Hence,
\begin{equation}
    \begin{split}
        q\del{\pmb{\hat{\mathcal{W}}}, \pmb{c}} &= \prod_l q\del{c_l^{-2}} \prod_i \prod_j q\del{\hat{w}_{ijl}} \\
        q\del{\hat{w}_{ijl}} &= \mathcal{N}\del{\mu_{ijl}, \sigma^2_{ijl}} \\
        q\del{c_l^{-2}} &=\mathcal{LN}\del{\mu_{c,l}, \sigma^2_{c,l}}.
    \end{split}
    \label{eq:ffpost}
\end{equation}

When inverting a hierarchical generative model over weights of artificial neural network, we exclusively apply stochastic black-box variational inference to the model variant with the regularised horseshoe prior. This choice is motivated by its documented superiority over the spike-and-slab prior, as established in \cite{ghosh2018structured}. We express the hierarchical prior in the non-centered parameterization as: 
\begin{equation*}
    \begin{split}
        a_{il}, b_{il} &\sim \Gamma\del{\frac{1}{2}, 1} \\
        \hat{a}_l, \hat{b}_l & \sim \Gamma\del{\frac{1}{2}, 1} \\
        \tau_{il} &= \sqrt{\frac{a_{il}}{b_{il}}} \\
        v_{l} &= \tau_0 \sqrt{\frac{\hat{a}_{l}}{\hat{b}_{l}}} \\
        \hat{w}_{ijl} &\sim \mathcal{N}\del{0, 1}\\
        w_{ijl} &= \gamma_{il} \hat{w}_{ijl}
    \end{split}
\end{equation*}
Note that the expressions above involve a reparameterization of Half-Cauchy distributed random variables as the square-root of the quotient of two Gamma distributed random variables, a strategy drawn from \cite{wand2011} (see \Cref{app:reparam} for additional details). Such a reparameterization of the Half-Cauchy ensures capturing of fat-tails in the posterior, even when leveraging a fully-factorised mean-field posterior approximation, as referenced in \cite{ghosh2018structured}.

For the fully-factorised mean-field approximation, the approximate posterior is portrayed as a composite of Normal and Log-Normal distributed random variables, expressed as:
\begin{equation*}
    \begin{split}
        q\del{\pmb{\hat{\mathcal{W}}}, \pmb{a}, \pmb{b}, \pmb{\hat{a}}, \pmb{\hat{b}}, \pmb{c}} &= \prod_l q\del{c_l^{-2}} q\del{\hat{a}_l}q\del{\hat{b}_l} \prod_i q\del{a_{il}}q\del{b_{il}} \prod_j q\del{\hat{w}_{ijl}} \\
        q\del{c_l} &= \mathcal{LN}\del{\mu_{c, l}, \sigma_{c, l}^2 } \\
        q\del{\hat{a}_l} &= \mathcal{LN}\del{\hat{\mu}_{a, l}, \hat{\sigma}_{a, l}^2 } \\
        q\del{\hat{b}_l} &= \mathcal{LN}\del{\hat{\mu}_{b, l}, \hat{\sigma}_{b, l}^2 } \\
        q\del{a_{il}} &= \mathcal{LN}\del{\mu_{a, il}, \sigma_{a, il}^2 } \\
        q\del{b_{il}} &= \mathcal{LN}\del{\mu_{b, il}, \sigma_{b, il}^2 } \\
        q\del{\hat{w}_{ijl}} &= \mathcal{N}\del{\mu_{w, ijl}, \sigma_{w, ijl}^2}
    \end{split}    
\end{equation*}

\subsection{Application of stochastic BMR to Bayesian neural networks}

To apply BMR to Bayesian neural networks, we commence by estimating an approximate posterior for the flat model, as detailed in
\cref{eq:flatprior}. To retain high computational efficiency, we pair BMR solely with the fully factorised approximate posterior, as presented in  \cref{eq:ffpost}. While it is feasible to use this method alongside the structured posterior \citep{ghosh2018structured}, it requires considerably more computationally intensive estimations of the reduced free energy. As shown below, we obtain satisfactory results with a fully factorised posterior. Therefore, we defer the exploration of BMR with a structured posterior to future endeavours.

Given a fully factorised approximate posterior, we can determine the change in variational free energy, $\Delta F$---after substituting the prior $\mathcal{N}\del{0, 1}$ with $\mathcal{N}\del{0, \theta_{ijl}^2}$ for the weight $\hat{w}_{ijl}$---as:
\begin{equation*}
    \begin{split}
        \Delta F\del{\theta_{ijl}} &= - \frac{1}{2}\ln \rho_{ijl}^2 - \frac{1}{2} \frac{\mu_{ijl}^2}{\sigma_{ijl}^2} \del{ 1 - \frac{\theta_{ijl}^2}{\rho_{ijl}^2} }  \\
        \rho_{ijl}^2 &= \theta_{ijl}^2  +  \sigma_{ijl}^2 - \theta_{ijl}^2 \sigma_{ijl}^2 
    \end{split}    
\end{equation*}

For the second hierarchical level of the approximate posterior, we aim to minimize the following form for the variational free energy:
\begin{equation}
    F = \sum_{l=1}^L E_{q\del{\pmb{\theta}_l}}\sbr{ - \sum_{i,j} \Delta F(\theta_{ijl}) + \ln \frac{ q\del{\pmb{\theta}_l} }{ p\del{\pmb{\theta}_l}} }
    \label{eq:bmrfe}
\end{equation}
This minimization is done with respect to $q\del{\pmb{\Theta}} = \prod_{l} q\del{\pmb{\theta}_{l}}$, the approximate posterior over hyper-parameters. Note the application of \cref{eq:loglkl} in substituting the marginal log-likelihood with the change in the variational free energy.

For the spike-and-slab prior we can write the following relation:
\begin{equation*}
    \begin{split}
        \pmb{\theta}_l &= \sbr{\pi_l, \lambda_{ijl} | \text{ for } i \in \cbr{1, \ldots, K_l}, \text{ and } j \in \cbr{1, \ldots, K_{l-1} + 1}} \\
        \theta_{ijl} &= \lambda_{ijl}
    \end{split}
\end{equation*}

Consequently, the approximate posterior at the second level of the hierarchy can be approximated as:
\begin{equation*}
    \begin{split}
        q\del{\pmb{\Theta}} &= \prod_{l} q\del{\pi_l} \prod_{ij} q\del{\lambda_{ijl}} \\
        q\del{\lambda_{ijl}} &= q_{ijl}^{\lambda_{ijl}} \del{1 - q_{ijl}}^{1 - \lambda_{ijl}} \\
        q\del{\pi_l} & = \mathcal{B}\del{\alpha_l, \beta_l}
    \end{split}
\end{equation*}

The iterative update to obtain the minimum of the simplified variational free energy (\cref{eq:bmrfe}) is then:
\begin{equation*}
\begin{split}
    q_{ijl}^{k+1} & = \frac{1}{1 + e^{- \sbr{\zeta^k_l - \Delta F\del{\lambda_{ijl}=0}} } }   \\
    \zeta^k_l &= \psi(\alpha^k_l) - \psi(\beta^k_l) \\
    \alpha^{k+1}_l & = \sum_{i,j} q^{k+1}_{ijl} + \alpha_0 \\
    \beta^{k+1}_l & = \sum_{i,j} \del{1 - q^{k+1}_{ijl}} + \beta_0 \\
\end{split}
\end{equation*}
Here, $\alpha^0_l = \alpha_0$, $\beta^0_l=\beta_0$, $ \Delta F\del{\lambda_{ijl}=0} = - \frac{1}{2} \sbr{ \ln \sigma_{ijl}^2 + \frac{\mu_{ijl}^2}{\sigma_{ijl}^2}}$, and $\psi\del{\cdot}$ refers to the digamma function. The efficiency of this inference scheme is remarkable, typically achieving convergence after a few iterations. In practice, we cap the maximum number of iterations at $k_{max}=4$.

Finally, we use the following pruning heuristics to eliminate model weights and sparsify network structure
\begin{equation*}
        \text{ if } q_{ijl}^{k_{max}} < \frac{1}{2} \text{, set } \hat{w}_{ijl} = 0.
\end{equation*}

To achieve the high sparsification rate presented in the next section, we adopt an iterative optimisation and pruning approach proposed in \cite{beckers2022principled}. We perform weight pruning at the beginning of each epoch (except the first one), and further optimisation for $500$ iterations, completing one epoch. In total, we apply iterative pruning and optimisation for fifty epochs in all examples below.

The complete implementation of stochastic BMR is available at an online repository \url{github.com/dimarkov/bmr4pml} with notebooks and scripts necessary to recreate all result figures. 

\section{Results}

In this section, we present the outcomes of our experiments and analyses conducted to evaluate the performance and efficiency of the stochastic Bayesian model reduction in the context of Bayesian sparsification of deep neural networks. Our results are structured to provide insights into the capabilities and advantages of our approach.

\subsection{Performance Comparison}

\begin{figure}[!h]
\centering
\includegraphics[width=\textwidth]{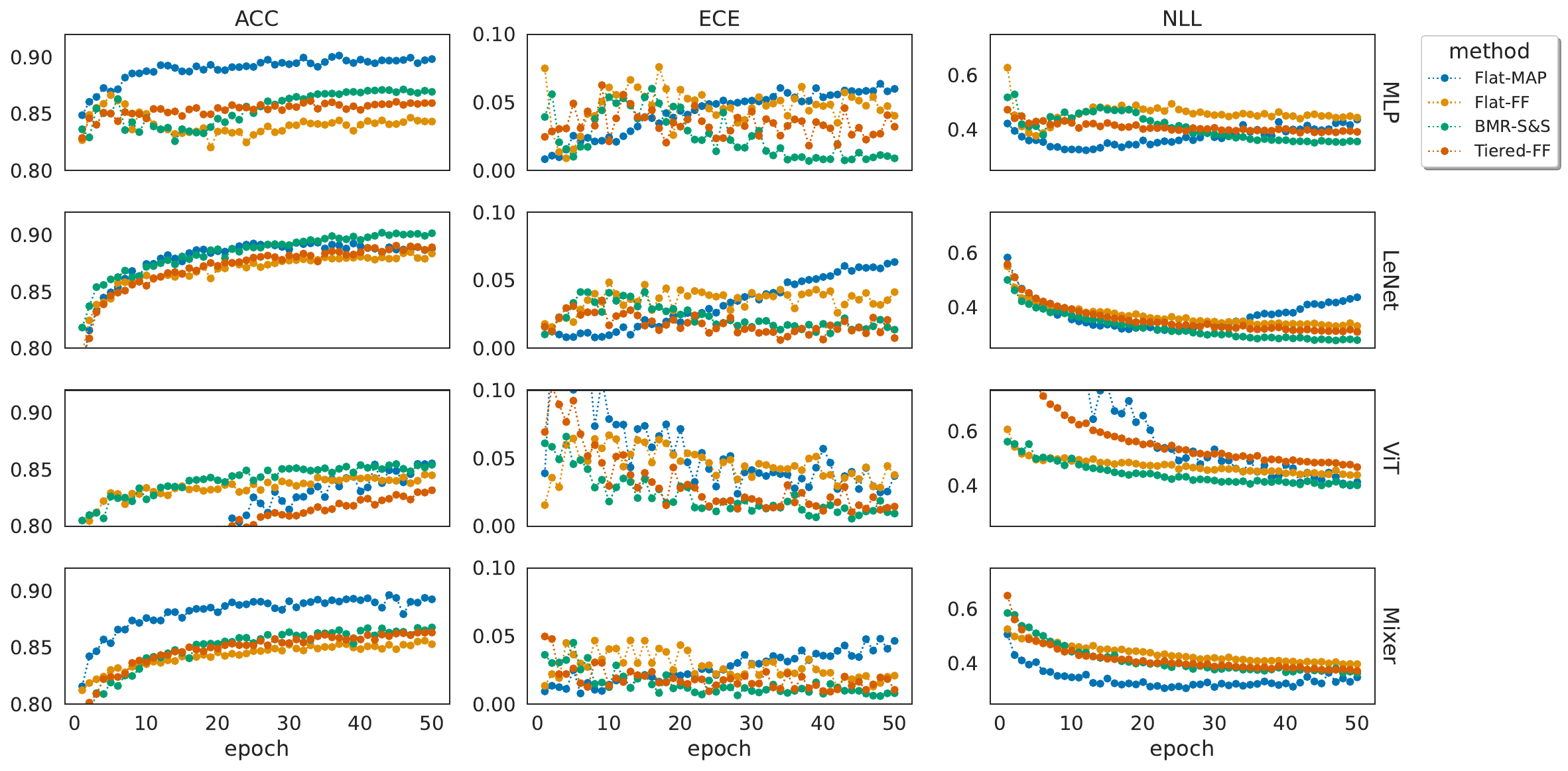}
\caption{\label{fig:Fig1} Classification performance comparison on FashoinMNIST dataset for different neuronal architectures and approximate inference schemes.}
\end{figure}

The training regimen used a batch size of $N_B = 128$ and the AdaBelief algorithm with learning rate set to $\alpha=10^{-3}$ in the case of the MAP estimate, $\alpha=5 \cdot 10^{-3}$ in the case of the mean-field methods, and $\alpha=10^{-2}$ in the case of stochastic BMR (the exponential decay rates were kept at default values $\beta_1=0.9$, and $\beta_2 = 0.999$). \Cref{fig:Fig1} charts the epoch-wise evolution of ACC, ECE, and NLL for each architecture, under five distinct approximate inference strategies: (i) Maximum a posteriori (MAP) estimate for the flat generative model, akin to traditional deep learning point estimates coupled with weight decay. (ii) A fully factorised posterior approximation for the flat generative model (Flat-FF). (iii) A fully factorised posterior approximation of the hierarchical generative model with a regularised horseshoe prior (Tiered-FF). (iv) The stochastic BMR algorithm augmented with a spike-and-slab prior (BMR-S\&S). Each epoch is defined by $500$ stochastic gradient steps, with each step randomly drawing $N_B$ data instances from the training pool. 

Interestingly, all approximate inference methods demonstrate comparable top-1 accuracy scores. However, the stochastic BMR method followed by the Tiered-FF approximatio (with a single exception), consistently resulted in the lowest ECE and NLL scores across the majority of DNN architectures and datasets (see \Cref{fig:Fig1-Sup1} for CIFAR10 dataset and \Cref{fig:Fig1-Sup2} for CIFAR100 dataset).The implicit reduction in model complexity suggests that---as anticipated---Stochastic BMR furnishes a model of the data that has the greatest evidence or marginal likelihood (not shown). In this setting, the NLL of the test data can be regarded as a proxy for (negative log) marginal likelihood.

\subsection{Learning of sparse representations}

\begin{figure}[!ht]
\centering
\includegraphics[width=\textwidth]{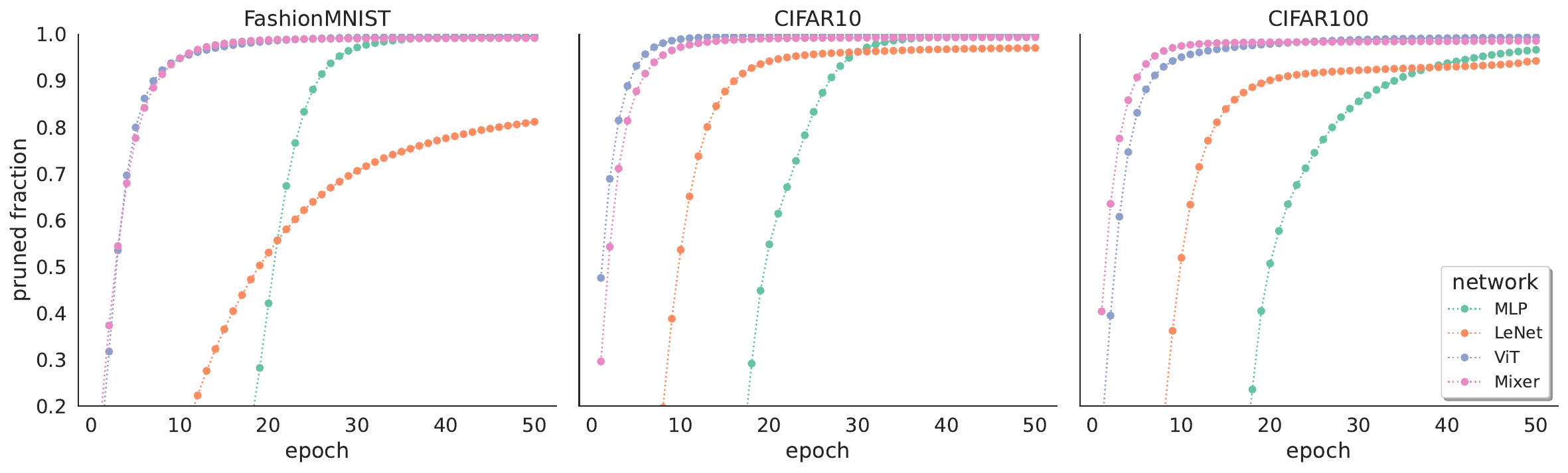}
\caption{\label{fig:Fig2}  Total fraction of pruned model parameters obtained with the stochastic BMR algorithm across different DNN architectures and datasets.}
\end{figure}

\Cref{fig:Fig2} depicts the fraction of pruned model parameters for different DNN architecures and datasets. It is noteworthy to observe the substantive sparsity achieved by the stochastic BMR algorithm. This sparsity is consistent across datasets and architectures, with the exception of the LeNet-5 structure when used for the FashionMNIST dataset, because by default LeNet-5 architecture is already sparse and contains relatively low-number of model weights (for other data sets we substantially increased the dimensionality of hidden layers as detailed in \Cref{app:dnns}). 

\begin{figure}[!h]
\centering
\includegraphics[width=\textwidth]{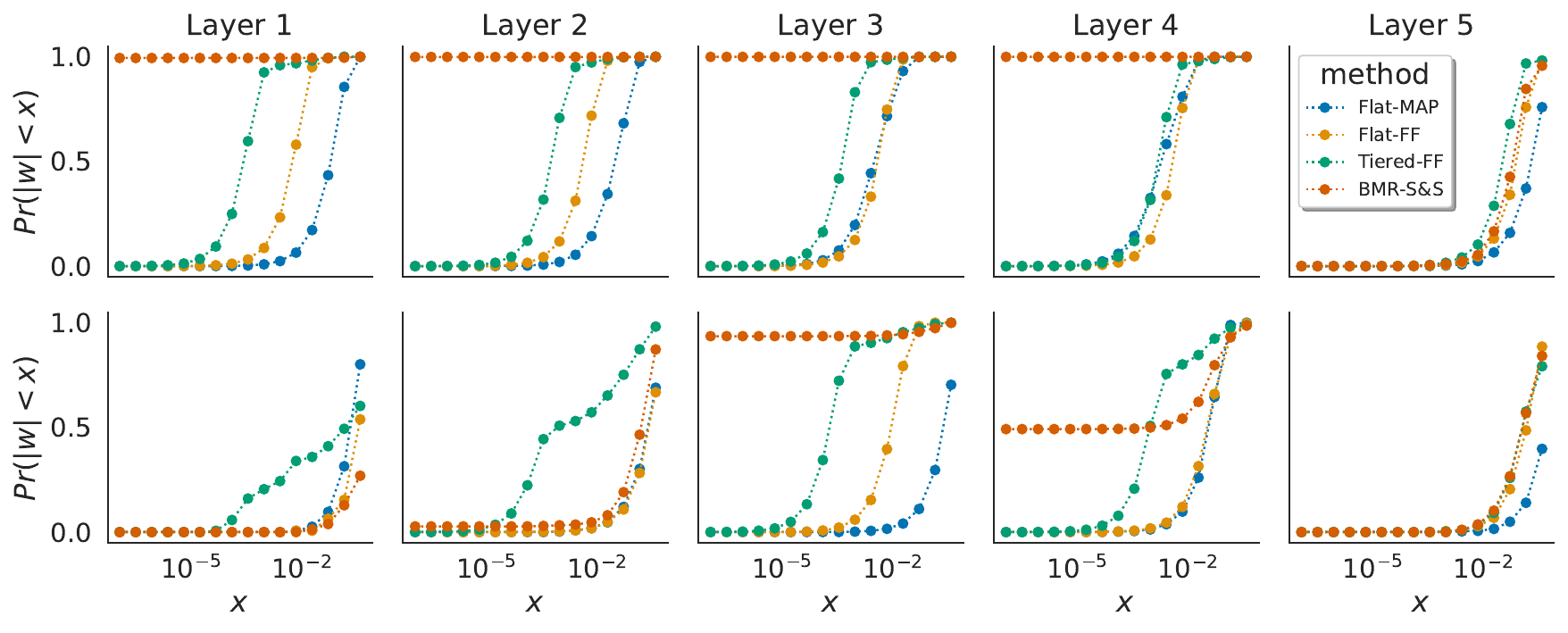}
\caption{\label{fig:Fig3}  Cumulative Distribution Function (CDF) of absolute posterior parameter expectations at different layers of MLP (top row), and LeNet architectures (bottom row). The y-axis represents the fraction of parameters with values less than or equal to the value on the x-axis.}
\end{figure}

To delve deeper into the pruning behavior across varying network depths, \Cref{fig:Fig3} presents a per-layer cumulative distribution function (CDF) for model parameters, highlighting the proportion of parameters whose absolute mean posterior estimate falls below a given threshold.
When juxtaposing the BMR CDF trajectories with those obtained from the Tiered-FF method (sparsification is induced by the regularised half-cauchy prior), it is evident that BMR furnishes more pronounced sparsification. This distinction is crucial, as the stochastic BMR not only matches or surpasses the performance of the Tiered-FF algorithm but also averages a $30\%$ faster stochastic gradient descent.

\begin{figure}[!h]
 \begin{subfigure}{.25\textwidth}
     \centering
     \includegraphics[width=\linewidth]{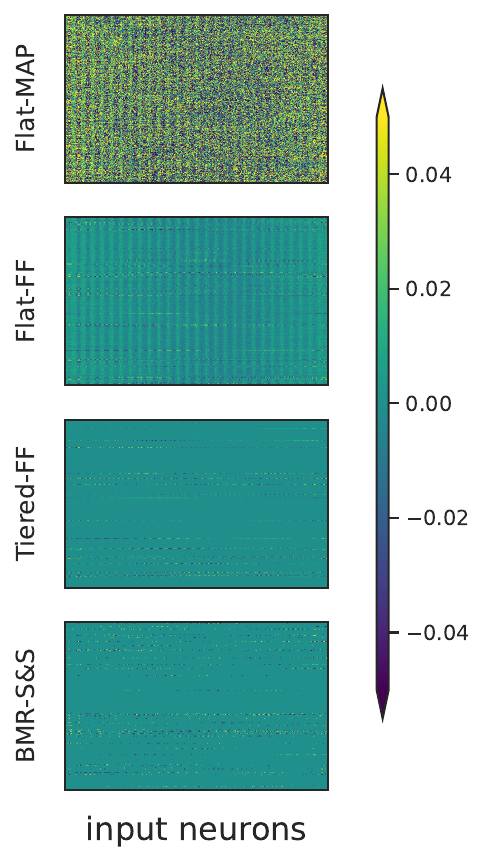}
     \caption{MLP}
     \label{fig:mlp}
 \end{subfigure}
 \begin{subfigure}{.75\textwidth}
     \centering
     \includegraphics[width=\linewidth]{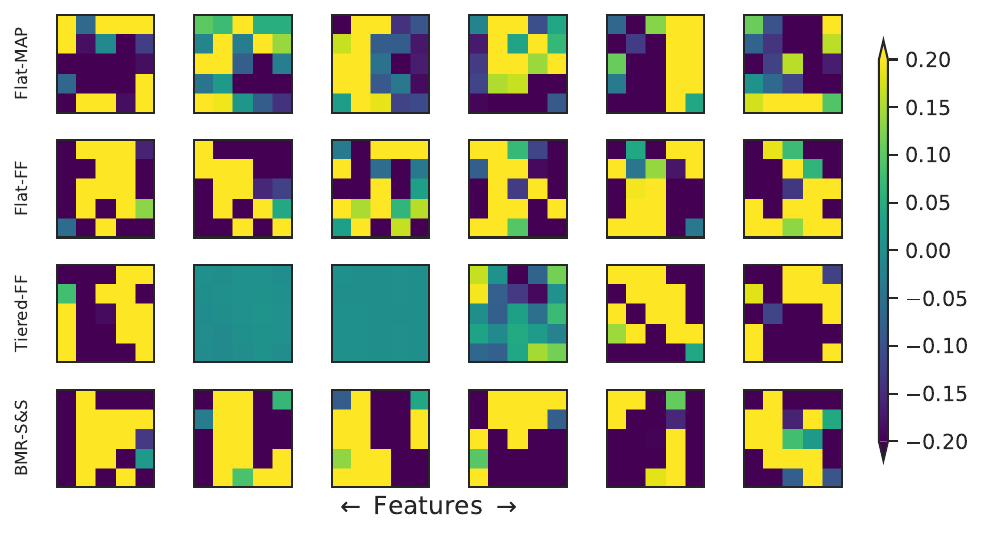}
     \caption{Lenet}
     \label{fig:lenet}
 \end{subfigure}
 \caption{Posterior expectations (color coded) over model parameters obtained using different approximate inference schemes at the first layer of (a) MLP architecture, and (b) LeNet architectures.}
 \label{fig:Fig4}
\end{figure}

To illustrate the structural learning variations among algorithms, \Cref{fig:Fig4} presents heatmaps of posterior expectations obtained using the four different methods. The \Cref{fig:Fig4} reveals subtle differences between inferred representations of the MLP and LeNet-5 architecture's input layers trained on the Fashion MNIST dataset. Divergent compression rates among the algorithms indicate inherent trade-offs between efficiency and performance. It is evident that the stochastic BMR strikes a balance between compression advantages and performance, as it is less prone to over-pruning as compared to the Tiered-FF method (two featured of the LeNet-5 input layer are effective removed - see \Cref{fig:Fig4}(b)). 

\section{Discussion}
In this study, we presented a novel algorithm---stochastic Bayesian model reduction---designed for an efficient Bayesian sparsification of deep neural networks. Our proposed method seamlessly integrates stochastic and black-box variational inference with Bayesian model reduction (BMR), a generalisation of the Savage-Dickey ratio. Through the stochastic BMR strategy, we enable iterative pruning of model parameters, relying on posterior estimates acquired from a straightforward variational mean-field approximation to the generative model. This model is characterized by Gaussian priors over individual parameters and layer-specific scale parameters. The result is an efficient pruning algorithm for which the computational demand of the pruning step is negligible compared to the direct stochastic black-box optimization of the full hierarchical model.

Over recent years, the Bayesian sparsification of neural networks has gained momentum, primarily driven by the spike-and-slab prior \cite{bai2020efficient,hubin2023variational,jantre2021layer,sun2022learning,ke2022optimization}. These works have showcased the remarkable sparsification capabilities inherent to such shrinkage priors. Nevertheless, when juxtaposed with the stochastic BMR algorithm, they often necessitate supplementary assumptions related to the approximate posterior. These assumptions, in turn, lead to a more computation-intensive model inversion. Moreover, in contrast to related approaches, the versatility of stochastic BMR allows its integration with more efficient optimization techniques, like variational Laplace \cite{daxberger2021laplace} and proximal-gradient methods \cite{khan2018fast}, provided the resulting approximate posterior in the form of a normal distribution is apt for the application at hand.

The insights obtained here pave the way for a deeper exploration of the potential applications of Bayesian model reduction across a wider array of architectures and tasks in probabilistic machine learning, such as audiovisual and natural language processing tasks. A more detailed fine tuning of the core dynamics of these algorithms, in terms of iterations steps, learning rates, and other free-parameters, might be the key to unveiling even more proficient Bayesian deep learning methodologies in the near future.

\acks{We thank Conor Heins, Magnus Koudahl, and Beren Millidge for valuable discussions during the initial stages of this work. SK acknowledges support by DFG TRR 265/1 (Project ID 402170461, B09) and Germany's Excellence Strategy---EXC 2050/1 (Project ID 390696704)---Cluster of Excellence ``Centre for Tactile Internet with Human-in-the-Loop'' (CeTI) of Technische Universität Dresden.
}

\newpage

\appendix
\section{}
\label{app:dnns}
For the simple multi-layer perceptron, we configure the architecture with five hidden layers, each comprising 400 neurons. The chosen activation function is the Swish activation function \citep{ramachandran2017searching}.

For the LeNet-5 architecture, we adhere to the original design, which includes three convolutional layers, average pooling following the initial two convolutional layers, and two linear layers. The activation function used is the hyperbolic tangent. The convolutional layers employ a kernel size of $5 \times 5$, while the average pooling uses a window of shape $2 \times 2$. For the FashionMNIST dataset, the feature counts of the convolutional layers are designated as (6, 16, 120), and the two linear layers have neuron counts of (84, 10). However, for the CIFAR10 and CIFAR100 datasets, we elevate the feature counts of the convolutional layers to (18, 48, 360), with linear layer neuron counts set to (256, 10) for CIFAR10 and (256, 100) for CIFAR100.

For the MlpMixer architecture we employ six layers and a patch resolution of $4 \times 4$. Across all datasets, we maintain constant values for hidden size ($C$), sequence length ($S$), MLP channel dimension ($D_C$), and MLP token dimension ($D_S$); specifically $C=256$, $S=64$, $D_C=512$ and $D_S=512$ for all datasets.

For the VisionTransformer architecture, we adopt a slightly modified version of the ViT-Tiny setup: we use six layers, eight heads for each attention block, an embedding dimension of 256, and a hidden dimension of 512. The patch resolution of $4 \times 4$ is consistent with the MlpMixer.
In both MlpMixer and VisionTransformer architectures, the GeLU activation function is used \citep{hendrycks2016gaussian}. 

For training using the maximum a posteriori estimate (Flat-MAP), dropout regularization, with dopout probability set to 0.2, is applied to all linear layers across all architectures, with the exception of the MlpMixer.

\section{}
\label{app:reparam}
In the centered parameterization of a generative model, Stochastic Variational Inference (SVI) with a fully factorized posterior yields a non-sparse solution, undermining the objective of employing shrinkage priors \citep{ghosh2019model}. Typically, this limitation is addressed by adopting the non-centered parameterization of the prior.

Consider the unique property of the half-Cauchy distribution: given $x \sim C^+(0, 1)$, and $z = b x$ the resulting probability distribution for $z$ is $z \sim C^+(0, b)$. Therefore, the non-centred parameterization is formulated as 
\begin{equation*}
    \begin{split}
        \hat{\tau}^l_i & \sim \mathcal{C}^+(0, 1) \\ 
        \hat{\lambda}^l_{ij} &\sim \mathcal{C}^+(0, 1) \\
        \hat{w}^l_{ij} & \sim \mathcal{N}\del{0, 1} \\
        \sbr{\gamma^l_{ij}}^2 & = \frac{\left[c^l \tau^l_0 \hat{\tau}^l_{i} \hat{\lambda}^l_{ij} \right]^2}{[c^{l}]^2 + \left[\tau_0^l \hat{\tau}^l_i \hat{\lambda}^l_{ij} \right]^2} \\
        w^l_{ij} &= \gamma^l_{ij} \hat{w}^l_{ij}
    \end{split}
\end{equation*}

However, while the half-Cauchy distribution is frequently chosen for sampling-based inference, it poses challenges in variational inference \citep{piironen2017sparsity}. Firstly, exponential family-based approximate posteriors (e.g., Gamma or log-Normal distributions) inadequately capture the half-Cauchy distribution's fat tails. Secondly, using a Cauchy approximating family for the posterior results in high variance gradients during stochastic variational inference \citep{ghosh2019model}. Hence, in the context of stochastic variational inference, the half-Cauchy distribution undergoes a reparameterization, as described in \citep{ghosh2018structured}:
\begin{equation*}
    x \sim \mathcal{C}^+(0, b) \equiv x = \sqrt{\frac{1}{u}}, u \sim \Gamma \del{\frac{1}{2}, \frac{1}{v}}, v \sim \Gamma \del{\frac{1}{2},b^2}
\end{equation*}
or, when represented in the non-centered parameterization:
\begin{equation}
    x = b \sqrt{\frac{v}{u}}, u \sim \Gamma \del{\frac{1}{2}, 1}, v \sim \Gamma\del{\frac{1}{2}, 1}
    \label{eq:rep_cauchy}
\end{equation}

\section{}

\beginsupplement

\begin{figure}[!h]
\centering
\includegraphics[width=\textwidth]{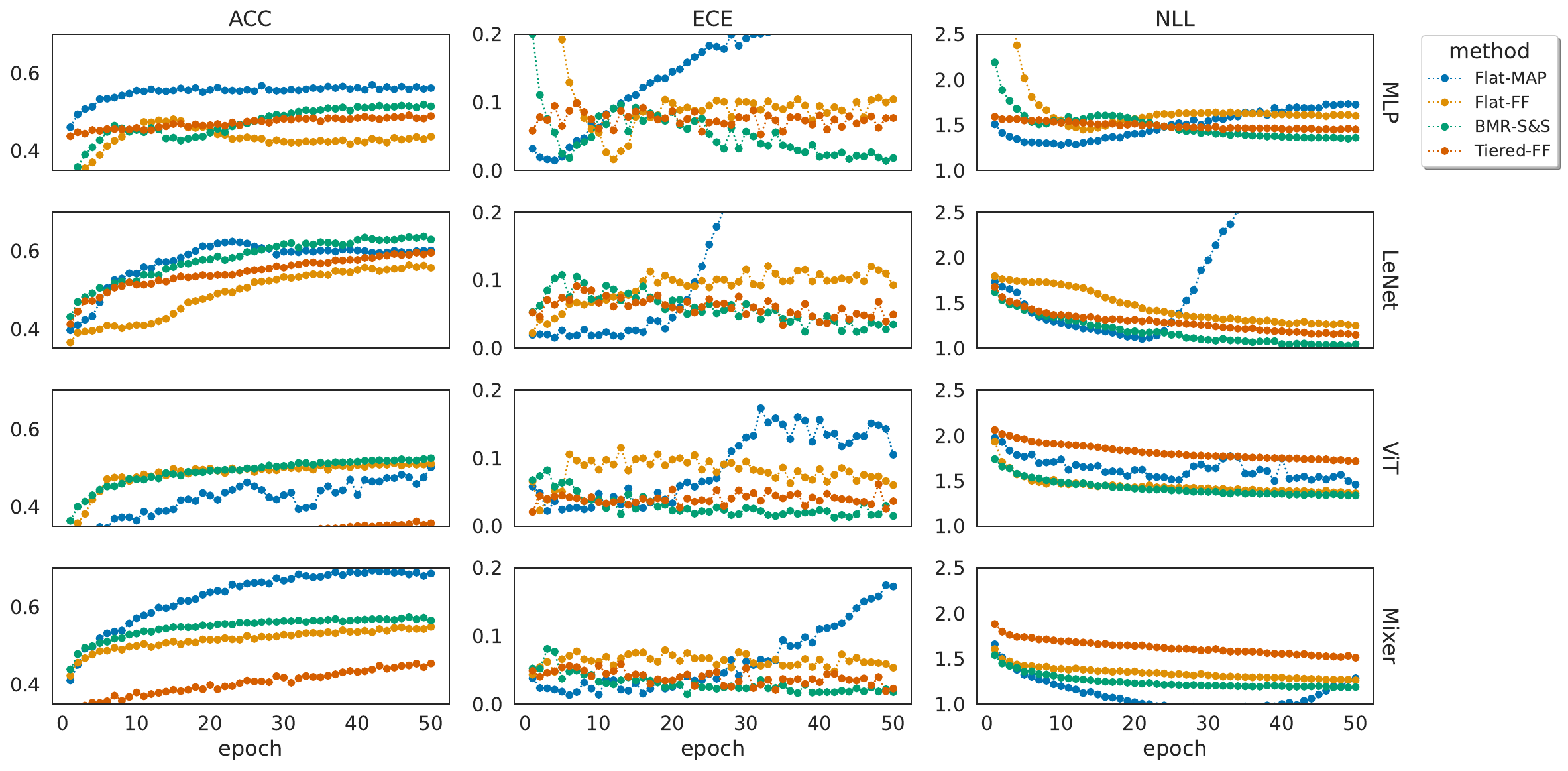}
\caption{\label{fig:Fig1-Sup1} Classification performance comparison on CIFAR10 dataset for different neuronal architectures and approximate inference schemes.}
\end{figure}

\begin{figure}[!h]
\centering
\includegraphics[width=\textwidth]{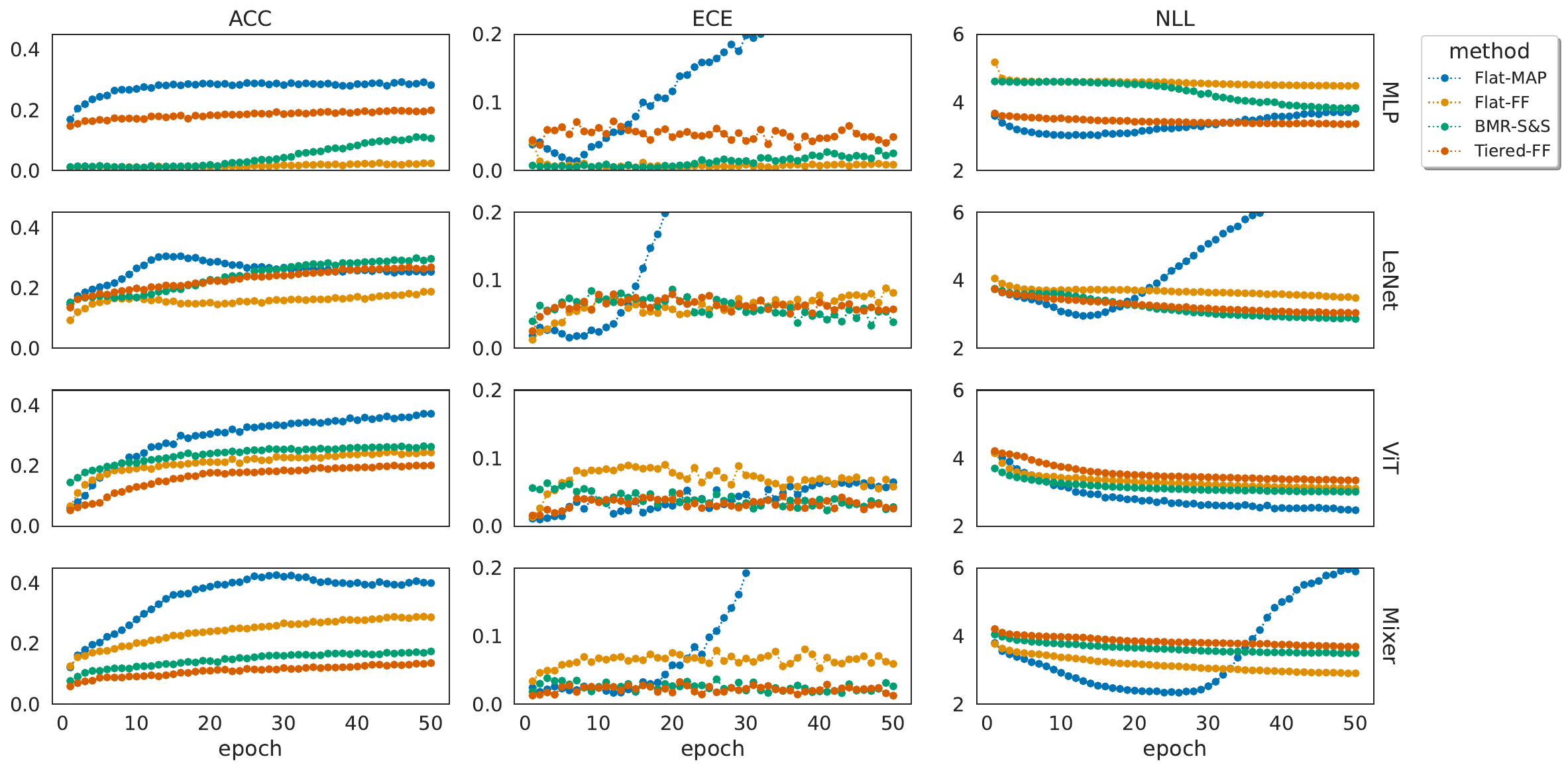}
\caption{\label{fig:Fig1-Sup2} Classification performance comparison on CIFAR100 dataset for different neuronal architectures and approximate inference schemes.}
\end{figure}

\bibliography{references}

\end{document}